\documentclass[12pt]{article}

\usepackage[letterpaper,left=1.25in,right=1.25in,top=1in,bottom=1in]{geometry}

\usepackage[T1]{fontenc}
\usepackage[utf8]{inputenc}
\usepackage{newtxtext,newtxmath} % Times-like text + math

\usepackage{setspace}
\usepackage{titlesec}
\titleformat{\section}{\bfseries\fontsize{16}{18}\selectfont}{\thesection}{1em}{}
\titlespacing*{\section}{0pt}{2\baselineskip}{1\baselineskip}

\titleformat{\subsection}{\bfseries\fontsize{14}{16}\selectfont}{\thesubsection}{1em}{}
\titlespacing*{\subsection}{0pt}{2\baselineskip}{1\baselineskip}

\titleformat{\subsubsection}{\bfseries\fontsize{12}{14}\selectfont}{\thesubsubsection}{1em}{}
\titlespacing*{\subsubsection}{0pt}{2\baselineskip}{1\baselineskip}

\usepackage{fancyhdr}
\usepackage{graphicx}
\usepackage{float}
\usepackage{booktabs}
\usepackage{caption}
\usepackage{pgfplots}
\usepackage{tikz}
\pgfplotsset{compat=1.17}

\usepackage[numbers]{natbib}

\usepackage[colorlinks=true,linkcolor=blue,citecolor=blue,urlcolor=blue]{hyperref}

\usepackage{array}
\usepackage{multirow}

\usepackage{amsmath}

\newcommand{\MICSTitle}[1]{%
  \begin{center}
    \vspace*{1.5in} % Title placed 1.5 inches below top
    {\fontsize{18}{20}\selectfont #1\par}
    \vspace{2\baselineskip} % 2 blank lines following
  \end{center}
}

\newcommand{\MICSAuthorBlock}[5]{%
  \begin{center}
    {\fontsize{14}{16}\selectfont #1\par} % Author name(s)
    {\fontsize{14}{16}\selectfont #2\par} % Department
    {\fontsize{14}{16}\selectfont #3\par} % Institution
    {\fontsize{14}{16}\selectfont #4\par} % Address (City, State Zip)
    {\fontsize{14}{16}\selectfont #5\par} % E-mail
    \vspace{2\baselineskip} % 2 blank lines following
  \end{center}
}

\newcommand{\MICSAbstract}[1]{%
  \begin{center}
    {\bfseries\fontsize{16}{18}\selectfont Abstract\par}
  \end{center}
  \vspace{\baselineskip} % 1 blank line following heading
  {\fontsize{12}{14}\selectfont
  #1\par}
}

\begin{document}

% ----------------------
% Title Page (NO number)
% ----------------------
\thispagestyle{empty}

\MICSTitle{Scribby: A Multi-Level LLM Framework for Semantic Video Analysis}

% If multiple authors, either list them on separate lines or combine as needed.
% Requirements: center, 14pt lines for name/department/institution/address/email.
\MICSAuthorBlock
  {Julian Abelarde, Hugo Garrido-Lestache Belinchon}
  {Department of Computer Science and Software Engineering}
  {Milwaukee School of Engineering}
  {Milwaukee, WI US}
  {\{abelardej,garrido-lestacheh\}@msoe.edu}

% Title page should consist of title and abstract only.
% Keep abstract 150 words or less, single spaced, blank line between paragraphs.
\newpage
\MICSAbstract{%
\begin{figure}[H]
\centering
\includegraphics[width=1\textwidth]{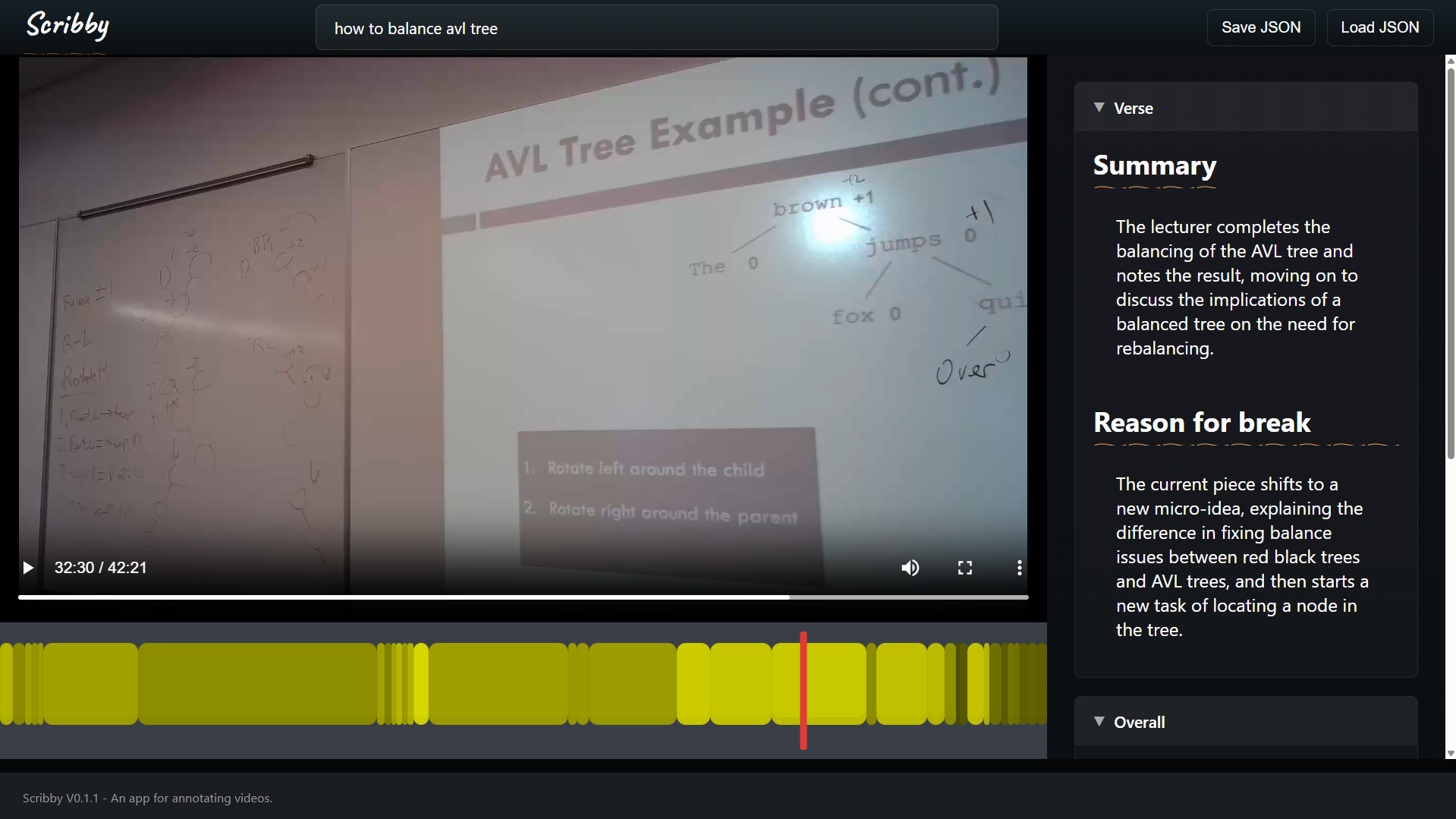}
\caption{Scribby user interface showing semantic verse clustering and interactive timeline. The system displays video chapters with semantic context and an interactive timeline where chapters are colored by cosine similarity to user-submitted queries, creating a relevance heatmap across the entire video.}
\label{fig:scribby_ui}
\end{figure}

As video content continues to expand across educational platforms, recorded lectures, and live-streamed entertainment, the need for efficient and structured analysis of long-form footage has increased \cite{1}. Although many existing AI programs provide a high-level video summary based on an AI-generated transcript \cite{2,3,4,5}, the approaches used are limited to a coarse overview and lack detailed analysis on a video's structure, thematic progression, and semantic relationships which are required to perform a comprehensive analysis of the video.

This papers proposes an LLM-based video summary framework that balances macro-level comprehension with micro-level semantic analysis \cite{6,12,13}. The first part of this process indexes the video on a micro level by (1) analyzing the video based on its entire transcript, (2) analyzing individual sentences of the transcript, and (3) grouping these sentences by semantic similarity by using LLM as a judge \cite{6,13}. Contextual continuity is retained during sentence-level processing by incorporating both the global transcript analysis and adjacent sentence information into each evaluation prompt.

After this indexing phase, semantically grouped sentence clusters--referred to in this paper as "verses"--are displayed in an interactive graphical timeline to visualize the semantic structure of the video, enabling users to identify thematic shifts and structural patterns \cite{15}. This semantic grouping is tested against various organized and unorganized web videos to ensure accuracy.

Furthermore, all verses are embedded in a vector-space representation to enable similarity-based search analogous to retrieval-augmented generation (RAG) systems \cite{14}. Similar testing methods to the graphical timeline are used to optimize accuracy and usability.

This framework establishes a foundation for visualizing video analysis tools like semantic chunking and matching by creating a heatmap measured on relevancy. Limitations and future expansions to the framework are also discussed.

\textbf{This work was made possible through ROSIE which provides freshman students unrestricted access to its computational power}.
}

% Body begins on a new page; page numbering starts at 1
\newpage
\setcounter{page}{1}
\pagestyle{fancy}

\section{Introduction}

The proliferation of video content on educational platforms, streaming services, and social media has created an unprecedented demand for efficient video understanding and summarization techniques. According to recent analysis, more than 500 hours of video content are uploaded to major platforms every minute \cite{1}, making manual analysis infeasible for most applications. This exponential growth is accompanied by increasing video lengths; tutorials, documentaries, and the rapidly growing niche of livestreaming produce videos that exceed several hours. Such an expansion requires robust automated summarization and navigation methods to help users efficiently access relevant content \cite{2}.

Current state-of-the-art video summarization approaches typically operate at a coarse granularity level \cite{3}. Most systems generate a single high-level summary of the entire video or extract a small set of key frames without capturing the fine-grained semantic structure that characterizes the evolution of topics throughout a video \cite{4}. Such approaches fail to preserve thematic progression and topical relationships that define a video's organizational structure. Users seeking specific information within long-form content must either rely on predefined chapter markers (if available) or engage in time-consuming manual navigation \cite{5}.

The challenge of video understanding is further complicated by content heterogeneity. Educational and structured content typically follows logical progressions with clear topical delineations, while unstructured content such as livestream recordings or creative videos may contain more fluid topic transitions \cite{3}. An effective video analysis framework must demonstrate robustness across both content types while maintaining semantic accuracy at the micro level.

This paper introduces Scribby, a multi-level LLM-based framework for semantic video analysis that addresses these limitations. Scribby combines macro-level video comprehension with sentence-level semantic analysis to create fine-grained, semantically-structured video representations \cite{6}. Rather than generating a single summary, Scribby segments videos into semantically coherent clusters called ``verses,'' preserving contextual relationships between content segments. These verses are then visualized in an interactive timeline and embedded in vector space to enable structural exploration and query-based retrieval.

\section{Related Work}

\subsection{Video Summarization: Foundations and Approaches}

Video summarization methods aim to condense full-length videos into concise synopses by selecting the most informative segments \cite{2}. Early and contemporary work emphasizes that modern techniques rely heavily on deep neural networks (CNNs, RNNs, attention mechanisms) to identify key frames or shots \cite{3,7}. These approaches often exploit multimodal data: in addition to video and audio, they leverage automatically transcribed speech (ASR transcripts), video captions, or other text metadata to guide summarization \cite{4}. Surveys by Apostolidis et al. demonstrate how leading summarizers incorporate transcripts and visual features together, achieving superior performance through multimodal fusion \cite{8}. Critically, most existing methods focus on coarse highlight detection and may overlook the fine-grained semantic structure necessary for detailed video comprehension \cite{3}.

\subsection{Transcription-Based Segmentation and Structure}

Automatically segmenting video transcripts into coherent topics has become essential, particularly for educational and long-form content \cite{5}. VT-SSum, a benchmark of 125K spoken-video transcript--summary pairs, demonstrates the value of transcript-based approaches on lecture and presentation videos \cite{9}. This work shows that slide content can weakly supervise extractive summaries, significantly improving performance on spoken-text summarization tasks.

Retkowski and Waibel's YTSEG framework \cite{10} defines the problem of ``smart chaptering''---predicting section boundaries and generating meaningful titles for video transcripts. They show that spoken content is more varied than written text, requiring hierarchical models such as MiniSeg to segment effectively \cite{10}. Similarly, Wang et al. provide a comprehensive study on lecture video fragmentation, classifying methods into boundary-based approaches (e.g., slide changes, pauses) and representation-based approaches (learned features of video chunks) \cite{11}. They introduce the MITFLD dataset and demonstrate that BiLSTM with self-supervised representations effectively detects lecture topic boundaries \cite{11}.

Scribby follows this general paradigm by fragmenting transcripts into micro-level representations, but emphasizes semantic coherence and interactivity rather than purely structural segmentation \cite{4}.

\subsection{LLM-Driven Video Analysis and Semantic Understanding}

Large language models have enabled deeper semantic understanding of video content. Lee et al. present LLMVS, a framework that generates frame captions using multimodal LLMs and scores frame importance through LLM-based contextual judgment \cite{6}. Their key insight---that LLMs encode vast general knowledge enabling semantic relevance judgment---directly informs Scribby's design. Similarly, Sugihara et al. propose a self-supervised pipeline where GPT-4 generates text summaries and selected frames are matched to summary concepts, demonstrating that powerful pre-trained LLMs can generate accurate summaries with minimal fine-tuning \cite{12}.

Tang et al.'s survey of Vid-LLMs categorizes LLM roles as summarizers, managers, and decoders \cite{13}. Using LLMs as summarizers involves prompting them to produce condensed descriptions from video-derived text, bringing ``open-ended multi-granularity reasoning'' and commonsense understanding to video analysis \cite{13}. Scribby leverages LLMs as semantic judges, evaluating sentence similarity and importance by combining global transcript context with local sentence relationships---similar to LLMVS's contextual relevance scoring \cite{6}.

\subsection{Video as Knowledge Base: RAG and Retrieval Approaches}

Retrieval-augmented generation approaches have recently been adapted for video understanding. VideoRAG extends RAG to long videos by constructing graph-based textual knowledge and multimodal embeddings, enabling retrieval of relevant segments across hours of content \cite{14}. The system implements hybrid retrieval combining graph-based structural reasoning with embedding-based semantic similarity \cite{14}.

Embedding-based retrieval allows efficient semantic search through vector space. Empirical results show that LLM-trained models significantly improve retrieval of key information from spoken transcripts \cite{9}. This approach enables applications such as query-based highlighting and timeline heatmaps of relevance \cite{14}. Scribby adopts embedding-based retrieval through Nomic-embed-text, visualizing semantic relevance as color-coded heatmaps.

\subsection{Interactive Visualization of Video Semantics}

Visualizing semantic structure is imperative for user comprehension. VideoForest, by Sun et al., introduces an interactive system using audience comments (``danmu'') to generate rich timeline visualizations \cite{15}. The system represents hierarchical relationships as tree structures (``scene forest'') with topic-level roots, allowing navigation by semantic topics. Evaluation with domain experts (animation professionals and academics) demonstrates effectiveness of forest-based metaphors for video exploration \cite{15}.

\begin{figure}[H]
\centering
\includegraphics[width=0.9\textwidth]{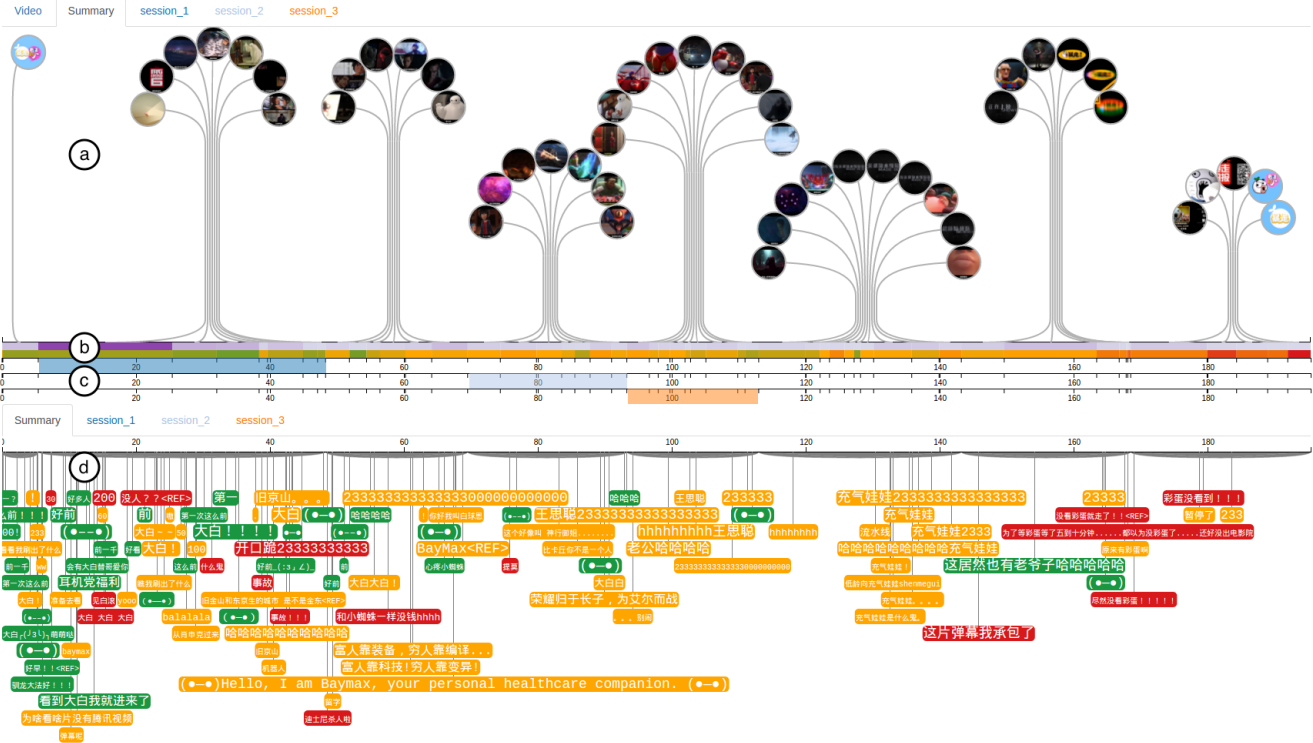}
\caption{VideoForest interactive visualization system. The interface displays (a) hierarchical scene trees representing video clusters, (b) a timeline ground showing video segments and metadata, and (c) annotated user comments (danmu posts) connected to relevant video moments. This design enables efficient semantic exploration and understanding of video structure through visual metaphors.}
\label{fig:videoforest}
\end{figure}

Other visualization systems (TimeSlice, LSTM-based skims) provide interactive GUIs for exploring long videos \cite{16}. Scribby's interactive timeline visualization similarly aims to make semantic structure visually explorable, though using a video timeline structure rather than forest metaphors. The qualitative evaluation methodology employed by VideoForest---gathering feedback from industry and academic experts---informs Scribby's evaluation approach \cite{15}.

\subsection{Integrating Multi-Modal Approaches: Synthesis of Current Work}

Current video analysis encompasses complementary but specialized approaches: transcription-based segmentation identifies structural boundaries \cite{10,11}, LLM-driven analysis enables semantic judgment \cite{6,12,13}, RAG-based retrieval supports knowledge access \cite{14}, and interactive visualization facilitates exploration \cite{15}. Scribby synthesizes these approaches into a unified framework. Micro-level sentence analysis provides fine-grained understanding \cite{6}, macro-level context ensures thematic continuity \cite{10}, LLM-based judgment enables semantic clustering \cite{13}, embedding-based retrieval enables search \cite{14}, and interactive visualization facilitates user exploration \cite{15}. This integration represents a holistic approach balancing computational efficiency with semantic richness, advancing beyond isolated solutions toward comprehensive video understanding.

\section{Methods}

\subsection{System Overview}

\begin{figure}[H]
\centering
\includegraphics[width=0.85\textwidth]{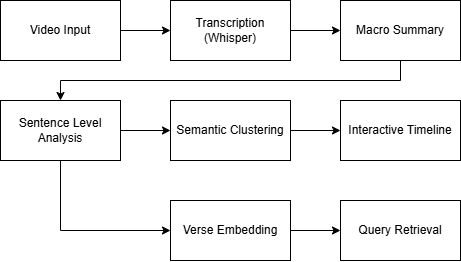}
\caption{Scribby system architecture showing the processing pipeline. Video input is transcribed to obtain sentence-level timestamps and a macro summary. Each sentence is semantically analyzed with contextual information and neural embeddings, and semantic boundaries are determined via LLM judgment, producing coherent verse clusters. The result is an interactive visualization of semantic verse clusters that enables query-based retrieval.}
\label{fig:scribby_overview}
\end{figure}

Scribby processes video content through three primary stages: (1) video transcription and macro-level summarization, (2) sentence-level semantic analysis and clustering, and (3) embedding-based indexing and interactive visualization. Figure \ref{fig:scribby_overview} illustrates the architecture.

\subsection{Video Transcription and Macro Summarization}

Video files are processed using FFmpeg to extract audio streams. Audio is transcribed using OpenAI's Whisper model \cite{6}, producing full transcripts with sentence-level timestamps and segmentation. This granularity serves as the foundation for micro-level analysis.

The complete transcript is then summarized using an LLM to generate high-level video overviews \cite{6}. This macro summary serves two functions: (1) it provides context for downstream sentence-level analysis, and (2) it enables the LLM to generate candidate chapter boundaries. The LLM identifies logical divisions in the transcript and generates descriptive summaries for proposed chapters \cite{10}.

\subsection{Sentence-Level Analysis and Contextual Embedding}

Rather than analyzing sentences in isolation, Scribby incorporates contextual information from multiple sources. For each transcript sentence, the system constructs a context including:

\begin{enumerate}
\item The sentence itself
\item The macro-level summary of the entire video
\item Summaries of adjacent sentences (preceding and following)
\item Historical context from previously-analyzed sentences within the same semantic cluster
\end{enumerate}

This multi-source context preservation ensures semantic judgments reflect each sentence's position within both local narrative (adjacent context) and global narrative (macro summary) \cite{6,13}. The LLM evaluates whether each sentence continues the current semantic cluster or initiates a new verse \cite{13}.

\subsection{Semantic Clustering via LLM Judgment}

Scribby employs LLMs as semantic judges for determining verse boundaries, leveraging contextual reasoning capabilities rather than purely embedding-based similarity thresholds \cite{13}. For each sentence, the system prompts the LLM with:

\begin{itemize}
\item The current verse being assembled
\item The sentence under evaluation
\item The macro-level summary providing global context
\item Explicit grouping criteria
\end{itemize}

The LLM responds with judgment of whether the sentence belongs to the current verse or initiates a new verse. This approach captures semantic nuances that distance-based metrics might miss, such as thematic shifts within sentences maintaining high lexical similarity \cite{6,13}.

\subsection{Verse Embedding and Vector-Space Representation}

Finalized verse summaries are embedded using Nomic-embed-text, a lightweight embedding model optimized for semantic search \cite{14}. Dimensionality is 768, standard for modern embedding models. These embeddings enable two capabilities:

\begin{enumerate}
\item \textbf{Heatmap Visualization:} Query embeddings are compared against verse embeddings using cosine similarity \cite{14}. Similarity scores (normalized to [0,1]) are mapped to color gradients, creating heatmaps of query relevance \cite{14}.

\item \textbf{Similarity-Based Search:} High-similarity verses can be combined to generate query-specific summaries, enabling content-aware extraction from long-form videos \cite{14}.
\end{enumerate}

\subsection{Interactive Visualization}

The visual interface displays the video timeline with color-coded verses. When a user submits a query, each verse's color intensity corresponds to its cosine similarity with the query embedding, enabling rapid visual identification of relevant sections \cite{15}.
\begin{figure}[H]
\centering
\includegraphics[width=1\textwidth]{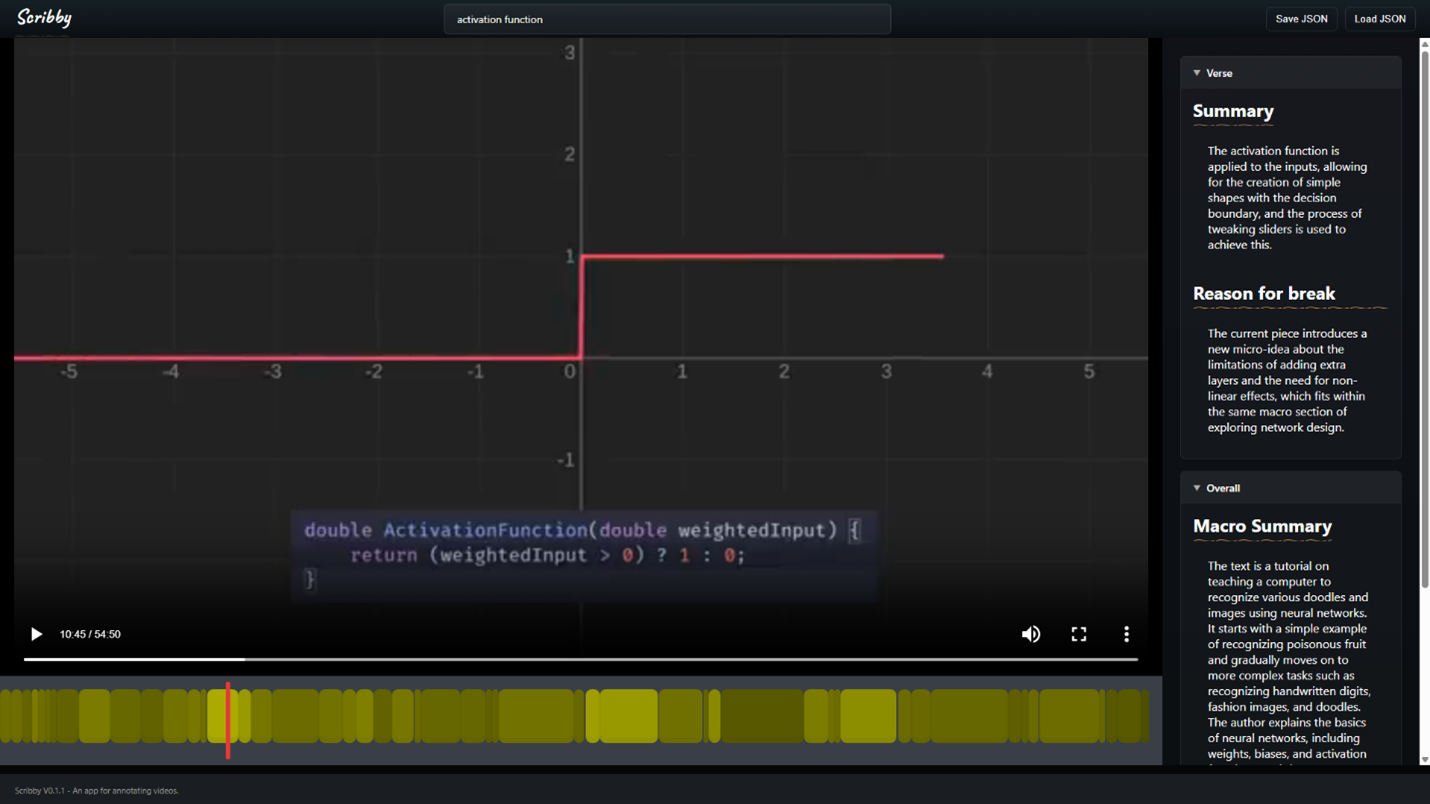}
\caption{Example of Scribby's UI in use. The user searched for clips related to "activation function" in a video about neural networks, and Scribby successfully highlighted the relevant segment, as demonstrated by the video clip visually emphasizing the activation function.}
\end{figure}

\section{Experiments}

\subsection{Experiment 1: Query Recall and Semantic Coherence}

\subsubsection{Methodology}

To evaluate correct identification of semantically related and unrelated content, we tested Scribby on \href{https://youtu.be/hfMk-kjRv4c?si=-nrtyQymtlxUn1oo}{``How to Create a Neural Network (and Train it to Identify Doodles)'' by Sebastian Lague}, a popular tutorial (2M+ views) with clear topical structure. We submitted queries spanning highly relevant to completely unrelated topics.

Query categories:
\begin{itemize}
\item \textbf{Related:} Terms directly addressed (``Neural network,'' ``Artificial intelligence'')
\item \textbf{Unrelated:} No relevance (``Marathon training,'' ``Construction management'')
\end{itemize}

Metrics computed across all verses:
\begin{itemize}
\item \textbf{Average Embedding Similarity:} Mean cosine similarity between query and verse embeddings
\item \textbf{Standard Deviation:} Variance across verses, indicating distributed versus isolated topic coverage
\end{itemize}

\subsubsection{Results}

\begin{table}[H]
\centering
\small
\begin{tabular}{lcc}
\toprule
\textbf{Query} & \textbf{Avg. Similarity} & \textbf{Std Dev} \\
\midrule
Neural network & 0.526 & 0.106 \\
Artificial intelligence & 0.453 & 0.062 \\
Marathon training & 0.413 & 0.060 \\
Construction management & 0.339 & 0.028 \\
\bottomrule
\end{tabular}
\caption{Query recall metrics. Related queries show higher average similarity and greater variance, indicating distributed topical coverage. Unrelated queries show consistently low similarity with minimal variance.}
\label{tab:query_recall}
\end{table}

The most relevant query (``Neural network'') achieved highest average similarity ($0.526$) and standard deviation ($0.106$), indicating thorough discussion in varying contexts. As relevance decreased, both metrics declined monotonically. The unrelated query produced lowest average similarity ($0.339$) and standard deviation ($0.028$), indicating uniform irrelevance. This relationship supports our hypothesis that relevant topics generate diverse similarity scores across contexts while irrelevant topics produce consistently low scores \cite{6,13}.

\subsection{Experiment 2: Alignment with Human-Defined Chapters}

\subsubsection{Methodology}

To evaluate semantic segmentation quality, we compared Scribby verses against YouTube chapters from the aforementioned \href{https://youtu.be/hfMk-kjRv4c?si=-nrtyQymtlxUn1oo}{``How to Create a Neural Network (and Train it to Identify Doodles)'' by Sebastian Lague} (21 chapters). YouTube chapters serve as ground truth representing expert semantic judgments. For each human chapter, we identified the closest Scribby verse and computed:

\begin{itemize}
\item \textbf{Time Offset:} Absolute difference between human and Scribby start times (seconds)
\item \textbf{Embedding Similarity:} Cosine similarity between human chapter name and Scribby verse summary
\item \textbf{Extra Verses:} Additional Scribby verses within each human chapter segment
\end{itemize}

We documented cases where LLM-generated summaries contained verbatim references to human chapter names (indicated as bolded in Table \ref{tab:chapter_alignment_1} and elaborated in Table \ref{tab:chapter_alignment_2}), providing strong evidence of semantic equivalence.

\subsubsection{Results}

\begin{table}[H]
\centering
\resizebox{\textwidth}{!}{
\setlength{\tabcolsep}{3pt}
\small
\begin{tabular}{|p{4cm}|c|c|c|c|c|}
\toprule
\textbf{Original Chapter Name} & \textbf{Original Chapter} & \textbf{Closest Chapter} & \textbf{Diff (s)} & \textbf{Summary-Title Sim} & \textbf{Extra Verses} \\
\midrule
Introduction & 0:00 & 0:00 & 0 & 0.523 & 6 \\
\textbf{The decision boundary} & 2:39 & 2:38 & 1 & 0.626 & 0 \\
\textbf{Weights} & 3:49 & 3:43 & 6 & 0.535 & 0 \\
Biases & 5:42 & 5:11 & 31 & 0.437 & 0 \\
Hidden layers & 6:45 & 6:38 & 7 & 0.488 & 0 \\
Programming the network & 7:45 & 7:41 & 4 & 0.551 & 2 \\
\textbf{Activation functions} & 9:57 & 9:46 & 11 & 0.655 & 2 \\
\textbf{Cost} & 12:42 & 12:49 & 7 & 0.561 & 0 \\
Gradient descent example & 15:07 & 15:03 & 4 & 0.648 & 3 \\
The cost landscape & 18:22 & 18:28 & 6 & 0.563 & 1 \\
Programming gradient descent & 19:55 & 19:52 & 3 & 0.489 & 0 \\
\textbf{It's learning! (slowly)} & 21:10 & 21:43 & 33 & 0.657 & 2 \\
Calculus example & 23:21 & 23:30 & 9 & 0.603 & 2 \\
The chain rule & 27:34 & 27:37 & 3 & 0.503 & 1 \\
Partial derivatives & 29:50 & 28:17 & 93 & 0.662 & 0 \\
Back propagation & 33:14 & 33:10 & 4 & 0.493 & 1 \\
Digit recognition & 39:25 & 39:21 & 4 & 0.620 & 4 \\
Drawing our own digits & 43:56 & 43:54 & 2 & 0.556 & 3 \\
\textbf{Fashion} & 47:37 & 47:34 & 3 & 0.599 & 0 \\
\textbf{Doodles} & 48:25 & 48:30 & 5 & 0.723 & 1 \\
The final challenge & 52:00 & 52:09 & 9 & 0.378 & 2 \\
\bottomrule
\end{tabular}
}
\caption{All 21 chapters from reference video with offsets when compared to the original video, embedding similarity scores, and extra verse counts. Mean offset: 11.29 seconds.}
\label{tab:chapter_alignment_1}
\end{table}

\begin{table}[H]
\centering
\resizebox{\textwidth}{!}{
\setlength{\tabcolsep}{2pt}
\small
\begin{tabular}{|l|p{6.2cm}|c|c|}
\toprule
\textbf{Chapter Name} & \textbf{Summary Excerpt} & \textbf{Summary-Title Sim} & \textbf{Avg-Title Sim} \\
\midrule
\textbf{The decision boundary} & … finding a \textbf{decision boundary} to classify fruit … & 0.626 & 0.447 \\
\textbf{Weights} & … outputs are calculated by multiplying the inputs with their respective \textbf{weights} and adding them together … & 0.535 & 0.410 \\
\textbf{Activation functions} & The \textbf{activation function} is applied to the inputs, allowing for the creation of simple shapes … & 0.655 & 0.431 \\
\textbf{Cost} & … find the optimal weights and biases that minimize the average \textbf{cost} … & 0.561 & 0.429 \\
\textbf{It's learning! (slowly)} & The mini batch technique speeds up \textbf{learning} iterations but introduces noise … & 0.657 & 0.510 \\
\textbf{Fashion} & The network achieves 89\% accuracy on the \textbf{fashion} dataset, outperforming human accuracy of 83\% … & 0.599 & 0.373 \\
\textbf{Doodles} & The author shifts focus to the goal of recognizing various \textbf{doodles} … & 0.723 & 0.404 \\
\bottomrule
\end{tabular}
}
\caption{Semantic correspondence for bolded chapters: LLM-generated verse summaries containing verbatim references to chapter names. \textbf{Summary Excerpt} shows exact matched phrases from generated summaries; \textbf{Summary-Title Sim} measures cosine similarity between verse summary and chapter name (mean: 0.654); \textbf{Avg-Title Sim} measures average cosine similarity across all verses to each chapter name (mean: 0.428).}
\label{tab:chapter_alignment_2}
\end{table}

Key findings:

\begin{enumerate}
\item \textbf{Temporal Accuracy:} Table \ref{tab:chapter_alignment_1} shows mean offset of 11.29 seconds across all 21 chapters, demonstrating accurate semantic boundary detection. Maximum offset (93 seconds) occurred in one case; most offsets were under 10 seconds. This validates that Scribby correctly identifies when semantic shifts occur within transcripts.

\item \textbf{Semantic Correspondence:} Table \ref{tab:chapter_alignment_2} displays the seven chapters where LLM-generated verse summaries explicitly contain verbatim references to human chapter names (indicated in bold in Table \ref{tab:chapter_alignment_1}). These items demonstrate particularly strong semantic alignment with mean Summary-Title Sim of $0.654$, compared to $0.535$ for non-bolded chapters. This validates that LLMs effectively identify nuanced semantic equivalences \cite{6,13}.

\item \textbf{Semantic Specificity vs. Global Baseline:} The disparity between Summary-Title Sim ($0.654$ mean) and Avg-Title Sim ($0.428$ mean) for strong semantic corresponding chapters is a 52.8\% higher similarity to their target chapters compared to average verses across the entire video. The Avg-Title Sim baseline represents the expected similarity if chapter names were randomly assigned; thus, this substantial gap (mean difference: 0.226) demonstrates recognition for highly relevant verses from general video content \cite{6,13}.

\item \textbf{Fine-Grained Segmentation:} Scribby generated 51 verses versus 21 human chapters (2.4x increase). While YouTube chapters represent high-level structural divisions, Scribby verses capture fine-grained semantic transitions valuable for detailed analysis and query-based retrieval \cite{10}.

\item \textbf{Extra Verses Distribution:} General chapters (e.g., ``Introduction'': 6 extra verses) contain multiple sub-topics, leading the LLM to segment further \cite{13}. Focused technical chapters (e.g., ``Weights'') remain cohesive with zero extra verses \cite{10}.
\end{enumerate}

\subsection{Experiment 3: Performance Across Content Types}

\subsubsection{Methodology}

A core design goal is structure-agnostic operation: consistent performance whether processing well-structured or exploratory content. We processed two videos of comparable length and similar technical topic:

\begin{itemize}
\item \textbf{Structured:} \href{https://youtu.be/_vqlIPDR2TU?si=Oy5Pam7Vl_sPEKK6}{``Coding Adventure: Making a Better Chess Bot''} (1:01:00) --- carefully-planned development log with clear progression
\item \textbf{Unstructured:} \href{https://youtu.be/IDx9iWqDwZE?si=HCqncR8rq7nvGQ-d}{``Chess Engine in Python - Part 12 - Greedy Algorithm and MinMax without recursion''} (52:47) --- unedited livestream with exploratory problem-solving
\end{itemize}

Measured metrics:
\begin{itemize}
\item Number of verses, average duration, variance in duration
\item Processing time: transcription, macro summary, sentence-level analysis
\end{itemize}

\subsubsection{Results}

\begin{table}[H]
\centering
\small
\begin{tabular}{lcc}
\toprule
\textbf{Metric} & \textbf{Structured (1:01:00)} & \textbf{Unstructured (52:47)} \\
\midrule
Number of verses & 61 & 50 \\
Average verse duration (s) & 58 & 63 \\
Std dev verse duration (s) & 66 & 75 \\
Maximum verse duration & 4:57 & 4:35 \\
\midrule
Transcription time & 2:58 & 2:32 \\
Macro summary generation & 0:05 & 0:10 \\
Sentence-level analysis & 9:32 & 6:07 \\
\textbf{Total processing time} & \textbf{12:35} & \textbf{8:49} \\
Processing time ratio & $0.206$ & $0.167$ \\
\bottomrule
\end{tabular}
\caption{Structural and temporal metrics comparing Scribby across content types, demonstrating consistency.}
\label{tab:content_types}
\end{table}

Results support structure-agnostic design:

\begin{enumerate}
\item \textbf{Structural Consistency:} Average verse duration differs by only 8.6\% (58s vs 63s), with comparable standard deviations (66s vs 75s). This indicates segmentation behavior is independent of content organization \cite{10,11}.

\item \textbf{Processing Efficiency:} Scribby processes videos at approximately 17--21\% of source duration (structured: 20.6\%, unstructured: 16.7\%). In contrast, manual expert review of technical video content requires approximately 100\% of source duration, as documented in our expert consultation.

As a consulted expert in video editing with nearly six years of professional YouTube video production experience, the lead author conducted focused time-motion analysis of typical video review workflows. The standard process involves watching source footage at 2--3$\times$ speed, supplemented by annotating observed segments. While faster playback nominally saves time, the concurrent annotation of meaningful observations (segment boundaries, key phrases, structural notes) introduces overhead that nearly offsets playback acceleration benefits. Across diverse content types, typical review time averages approximately 100\% of source video duration, with observable variance: highly repetitive or structural content can be reviewed in approximately 75--85\% of source time, while complex, densely-packed technical content typically requires 115--125\% of source time to adequately capture all relevant transitions and discussions. This 100\% baseline therefore represents a robust and conservative estimate of expert manual review time \cite{17}.

Scribby thus achieves approximately a 5--6x efficiency gain compared to manual expert review. Both videos processed end-to-end in under 13 minutes, demonstrating practical scalability for educational and professional video analysis \cite{11}.
\end{enumerate}

\begin{figure}[H]
\centering
\includegraphics[width=0.95\textwidth]{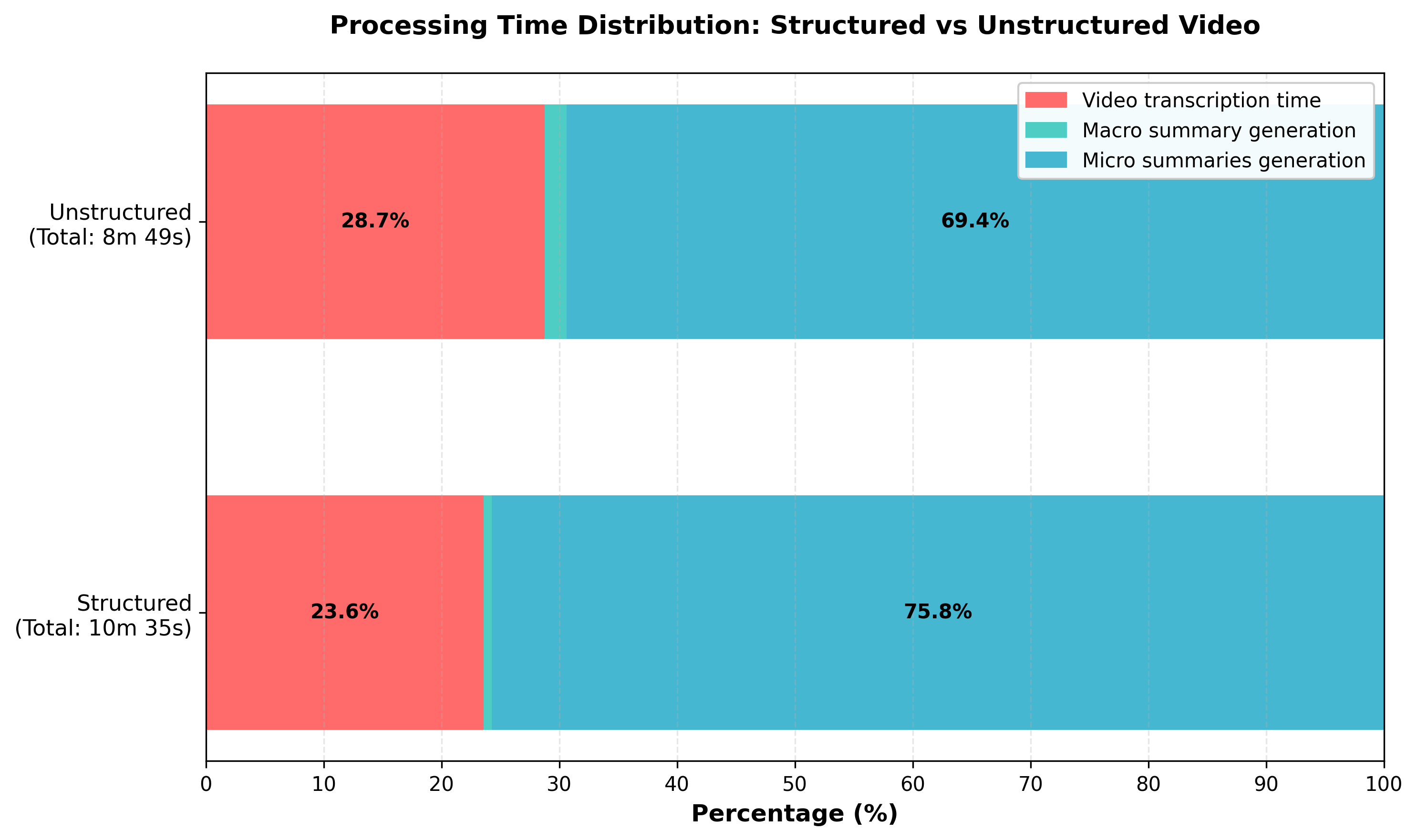}
\caption{Processing time distribution across pipeline stages for structured vs. unstructured videos. Top: Unstructured video (Scribby total: 8m 49s). Bottom: Structured video (Scribby total: 12m 35s). Both charts show the percentage breakdown of transcription time (red), macro summary generation (light blue/teal), and micro summaries generation (dark blue/teal) stages, demonstrating consistent processing ratios across content types and structure-agnostic performance.}
\label{fig:processing_chart}
\end{figure}

Processing time visualization is presented in Figure \ref{fig:processing_chart}. The chart displays the distribution of computational work across pipeline stages (video transcription, macro summary generation, and micro summaries generation) as percentages of total processing time. 

Key observations: the unstructured video process took 16.7\% of the source video length (28.7\% transcription, 1.1\% macro summary, 69.4\% sentence-level analysis), while the structured video process took 20.6\% of the source video length (23.6\% transcription, 0.4\% macro summary, 75.8\% sentence-level analysis). This distribution demonstrates that sentence-level semantic analysis dominates computational cost while processing composition remains consistent across content types \cite{6,11}.

\section{Discussion}

Experimental results demonstrate that Scribby effectively balances macro-level and micro-level analysis. Query recall (Experiment 1) validates correct semantic distinction between relevant and irrelevant content. Chapter alignment (Experiment 2) shows strong correspondence with human expert judgment (11.29 second mean offset, 0.654 similarity for semantically validated cases). Content type analysis (Experiment 3) confirms consistent operation across structured and unstructured videos \cite{6,10,11,13}.

However, important limitations exist. Embedding similarity metrics, while generally effective, do not capture complete semantic equivalence \cite{14}. Unbolded chapters in Experiment 2 showed lower similarity despite capturing correct concepts, suggesting that embedding-based metrics may underestimate semantic correspondence when summary language diverges from chapter naming conventions \cite{14}. Domain-specific terminology may receive lower similarity scores despite capturing creator intent.

Additionally, LLM-based judgment introduces potential inconsistencies \cite{6,13}. While LLMs demonstrate robustness across domains, prompt formulation, model state, and inherent stochasticity may produce varying segmentation decisions. Mission-critical applications (e.g., medical lecture analysis) require calibration against ground truth \cite{13}.

Computational overhead remains non-negligible. Sentence-level analysis consumes 50--60\% of total time, reflecting repeated LLM inference costs \cite{6}. Real-time processing would require optimization through batching or distillation to smaller models \cite{13}.

\subsection{Limitations and Future Work}

Important limitations:

\begin{enumerate}
\item \textbf{Embedding Similarity Limitations:} Cosine similarity conflates multiple meaning dimensions \cite{14}. Two summaries with high similarity might share vocabulary while differing in intent, relevant for specialized applications.

\item \textbf{Limited Evaluation Scope:} Evaluation focused on technical tutorials with clear structure. Diverse content types (news, entertainment, creative work) would strengthen generalizability claims \cite{10,11}.

\item \textbf{LLM Dependency:} Framework performance depends on LLM capability. Quality degradation or availability issues directly impact performance \cite{6,13}.

\item \textbf{Query Formulation Burden:} Users must formulate effective queries \cite{14}. Heatmap visualization aids exploration but could benefit from automatic query suggestion and reformulation.
\end{enumerate}

Future work directions:

\begin{enumerate}
\item \textbf{Hybrid Similarity Metrics:} Combine embedding-based and structure-based metrics capturing multiple semantic dimensions \cite{14}.

\item \textbf{Multi-Modal Analysis:} Incorporate visual information (scene detection, OCR, object recognition) for improved segmentation where visual transitions are semantically significant \cite{15}.

\item \textbf{User-Guided LLM Prompting:} Enable users to provide initial understanding or context about video topics, allowing the LLM to refine its semantic judgment criteria based on user-specified focus areas. This would increase accuracy for domain-specific analysis where users possess prior knowledge.

\item \textbf{Qualitative Evaluation:} Conduct formal user studies with professional video creators and content consumers to assess practical utility and interface improvements \cite{15}.

\item \textbf{Real-Time Processing:} Optimize the pipeline for livestream content processing \cite{11}.

\item \textbf{Scalability:} Explore distributed processing architectures and edge deployment for large-scale corpora \cite{11}.

\item \textbf{Automatic Video Editing:} Combine relevant video clips to summarize a long, unorganized video into a digestible overview. This can also speed up the workflow of video editors in film-making, content creation, and related fields. 
\end{enumerate}

\section{Conclusion}

This paper introduces Scribby, a multi-level LLM-based framework for semantic video analysis combining sentence-level LLM judgment with contextual embedding and interactive visualization. Experimental evaluation demonstrates that Scribby correctly discriminates between related and unrelated topics, achieves strong correspondence with human expert chapter boundaries (11.29-second mean offset), and operates consistently across both structured and unstructured content. The framework achieves approximately 5--6x faster processing compared to manual expert review while generating fine-grained semantic verse clusters that preserve thematic structure throughout video content.

The results establish Scribby as a foundation for advanced video analysis and retrieval systems, with immediate applications in query-based highlighting, semantic summarization, and interactive exploration. Future work will extend evaluation to diverse content types, incorporate multi-modal analysis, enable user-guided LLM prompting, and optimize for real-time processing. Scribby demonstrates that LLM-based semantic judgment combined with vector-space retrieval provides a promising paradigm for comprehensive video understanding.

\section{Ackowledgements}
\textbf{This work was made possible through ROSIE which provides freshman students unrestricted access to its computational power}. This enabled large-scale LLM experimentation without costly API calls; in particular, ROSIE provided the necessary compute capacity to run and evaluate large language models such as LLaMA 70B that would be infeasible to deploy on school hardware. The availability of these resources significantly accelerated development, iteration, and empirical validation throughout the project.

\bibliographystyle{plainnat}

\end{document}